\documentclass[letterpaper, 10 pt, conference]{ieeeconf}  

\IEEEoverridecommandlockouts                              

\usepackage{lettrine}
\usepackage{tikz}
\usepackage{amsfonts,amssymb}
\usepackage{amsfonts}
\usepackage{amsmath}
\usepackage[ruled,linesnumbered]{algorithm2e}
\usepackage{algpseudocode}
\usepackage{amsmath}
\usepackage{graphics}
\usepackage{epsfig}
\usepackage{color}
\usepackage{verbatim}
\usepackage{multirow}
\usepackage{color}
\usepackage{colortbl}

\title{\LARGE \bf
A Learning-Driven Framework with Spatial Optimization For Surgical Suture Thread Reconstruction and Autonomous Grasping Under Multiple Topologies and Environmental Noises}

\author{Bo Lu$^{1}$, Wei Chen$^{1}$, Yue-Ming Jin$^{{2}}$, Dandan Zhang$^{3}$, Qi Dou$^{{2}}$, \textit{Member, IEEE}, Henry K. Chu$^{4}$, \\ \textit{Member, IEEE}, Pheng-Ann Heng$^{{2}}$, \textit{Senior Member, IEEE}, and Yun-Hui Liu$^{{1}}$, \textit{Fellow, IEEE}
\thanks{This work is supported in part of the HK RGC under T42-409/18-R and 14202918, in part by project 4750352 of the CUHK-SJTU Joint Research Fund, and in part by the VC Fund 4930745 of the T Stone Robotics Institute.}
\thanks{$^{1}$B. Lu, W. Chen and Y. H. Liu are with the T stone Robotics Institute, The Department of Mechanical and Automation Engineering, The Chinese University of Hong Kong. Y. H. Liu is the Director of the T Stone Robotics Institute.
	{Contact email: \tt\small: bolu@cuhk.edu.hk}.}
\thanks{$^{2}$Y. M. Jin, Q. Dou and P. A. Heng are with The Department of Computer Science and Engineering, The Chinese University of Hong Kong.}%
\thanks{$^{3}$D. Zhang is with The Hamlyn Centre for Robotic Surgery, Imperial College London, UK.}%
\thanks{$^{4}$H. K. Chu is with The Department of Mechanical Engineering, The Kong Kong Polytechnic University, Hong Kong.}%
}

\begin{document}

\maketitle
\thispagestyle{empty}
\pagestyle{empty}

\begin{abstract}
Surgical knot tying is one of the most fundamental and important procedures in surgery, and a high-quality knot can significantly benefit the postoperative recovery of the patient. However, a longtime operation may easily cause fatigue to surgeons, especially during the tedious wound closure task. 
In this paper, we present a vision-based method to automate the suture thread grasping, which is a sub-task in surgical knot tying and an intermediate step between the stitching and looping manipulations. 
To achieve this goal, the acquisition of a suture's three-dimensional (3D) information is critical. Towards this objective, we adopt a transfer-learning strategy first to fine-tune a pre-trained model by learning the information from large legacy surgical data and images obtained by the on-site equipment. Thus, a robust suture segmentation can be achieved regardless of inherent environment noises.
We further leverage a searching strategy with termination policies for a suture's sequence inference based on the analysis of multiple topologies. Exact results of the pixel-level sequence along a suture can be obtained, and they can be further applied for a 3D shape reconstruction using our optimized shortest path approach.
The grasping point considering the suturing criterion can be ultimately acquired.
Experiments regarding the suture 2D segmentation and ordering sequence inference under environmental noises were extensively evaluated. Results related to the automated grasping operation were demonstrated by simulations in V-REP and by robot experiments using Universal Robot (UR) together with the da Vinci Research Kit (dVRK) adopting our learning-driven framework. 
\end{abstract}

\section{INTRODUCTION}
\lettrine[lines=2]{R}{aises} of modern techniques in the last few decades prompt a rapid development and iteration of robotic system \cite{Taylor_Russell_H}. With the aids of sophisticated designs \cite{J_Dargahi}, advanced materials \cite{C_W_Lee}, and computer-integrated technologies \cite{R_H_Taylor_2}, robot-assisted surgery (RAS) came into use in response to requirements of higher dexterity, accuracy and efficacy in clinical operations \cite{D_B_Camarillo}, such as in laparoscopic interventions of prostatectomy  and hysterectomy in the abdominal cavity \cite{H_Reich}.
Nowadays, the prosperous growths of medical robots not only benefit practical operations of surgeons, but also alter their manipulating paradigms and significantly extend modern surgical capabilities. 

Surgical suturing/knot tying for cuticle wound closure \cite{lu_bo_1} is a fundamental and important task, yet it remains a challenging operation in minimally invasive surgery (MIS), such as in endoscopic vascular anastomosis and transanal total mesorectal excision (taTME) for low rectal cancer \cite{E_C_Mclemore}.
Rather than conducting it manually, this task can now be operated remotely and ergonomically via a 3-D visualization and EndoWrist equipment \cite{H_P_Jaydeep} using the cutting-edge master-slave surgical robot (e.g. da Vinci robotic system).
However, a longtime surgery requires intensive attentions and easily brings fatigue to surgeons. This may result in low-quality wound closure or even lead to extra trauma to patient.
To release burdens from surgeons, immense efforts have been leveraged to automate each standard step in surgical knot tying task \cite{lu_bo_1}\cite{lu_bo_2}, promoting the development of vision-based automated robotic surgery.
\begin{figure}[t]
\centering
\includegraphics[width=0.48\textwidth]{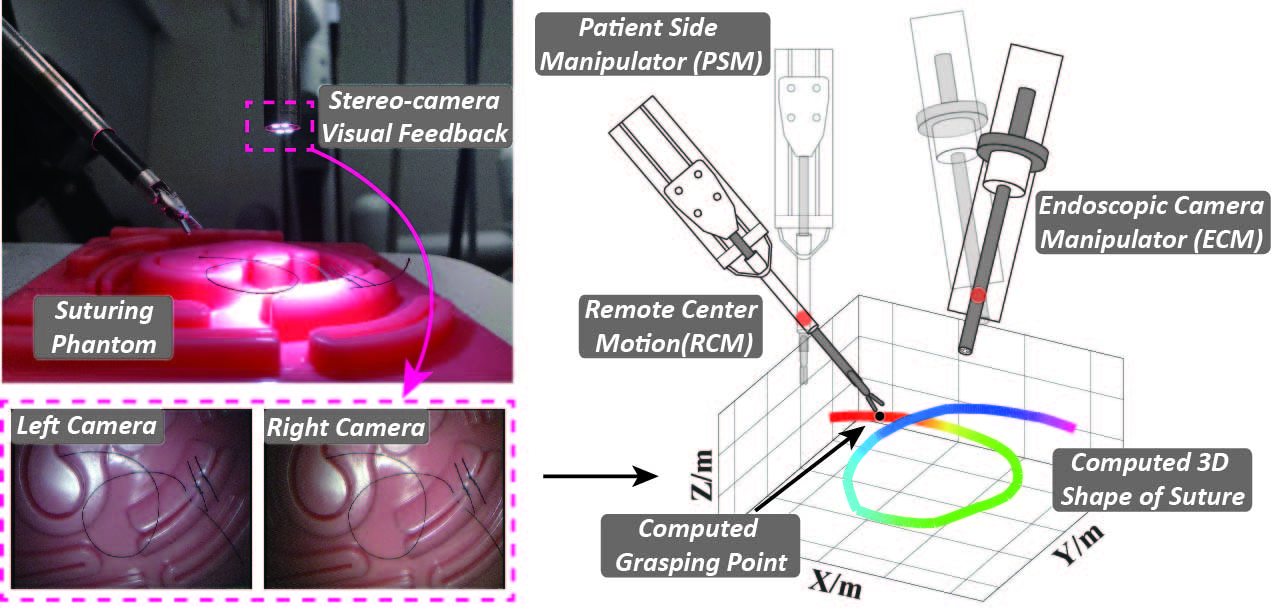}
	\caption{With calibrated stereo visions, 3D coordinates and the grasping point of an arbitrary orientated suture can be obtained using our framework, thereby automating a vision-based suture grasping manipulation.}
	\label{figure_intriduction_illustration}
\vspace{-0.5cm}
\end{figure}

To initialize the surgical knot tying, extensive investigations have been put on automated stitching/piercing tasks, covering the research areas of
collaborative hardware developments for robotic suturing \cite{Siddarth_sen}\cite{Hu_yang}\cite{S_Leonard}, 
semantic segmentation of surgical tools \cite{I_Laina}, 
task planning and execution via learn-by-demonstrations \cite{Huang_bidan} and supervised learning \cite{A_Shademan}, 
interaction analysis between rigid instrument and soft tissue \cite{F_Alambeigi}\cite{David_navarro}.
All these pioneering studies provided potential solutions to knotty challenges in automated suturing task \cite{Yang_guangzhong}.
However, the wounds were all anastomosed by continuous sutures in \cite{Siddarth_sen}\cite{S_Leonard}\cite{A_Shademan}, and it may result in the loose of the entire suturing if one stitch breaks, thereby harming the rehabilitation of patient.

To overcome this problem, we adopt the interrupted suturing technique, in which suture knots should be generated sequentially along a surgical wound. 
Based on pre-mentioned works, suture can be automatically stitched through soft tissues, setting a solid foundation for the coming knot tying task. 
In \cite{lu_bo_1}, an optimal suture's looping trajectory was generated and a vision-based method was proposed for the automated knot tying task. Besides, a robot can autonomously learn an online trajectory for a suture's looping task, and further, a force controller was integrated into the whole system to dynamically monitor motions of robot \cite{T_Osa_tase}. 
However, the thread should be cut off when using the interrupted suturing \cite{lu_bo_tmech}, and thereby, the grasping of a cutting thread is the key to realizing a fully automated robotic knot tying task.

To address the critical issues mentioned above, researchers put their attentions on suture segmentation, topological analysis, 3D reconstruction and relevant manipulations.
Padoy et al. \cite{N_Padoy_IROS_2011} presented a method for tracking deformable 3D suture based on discrete Markov random field optimization and a non-uniform rational B-spline (NURBS).
Later on, Jackson et al. \cite{R_C_Jackson_tase_2018} adopted the NURBS curve to obtain 3D coordinates of a suture by minimizing its image matching energy in a stereo-camera system.
Besides, Gu et al. \cite{Gu_yun_2018_IROS} proposed a joint feature learning framework for suture's detection by semi-supervised and unsupervised learning domain adaptation.
In addition, Lui \cite{W_H_Lui} used particle filter and learning algorithm to infer and untangle a rope's knot with the data acquired from an RGB-D camera. 
However, these existing works only focused on 2D image semantic segmentation \cite{Gu_yun_2018_IROS}, or didn't examine the suture's reconstruction with multiple shapes under complex environments \cite{N_Padoy_IROS_2011}\cite{R_C_Jackson_tase_2018}. The relevant manipulations are only limited to macro-size cables \cite{W_H_Lui}\cite{Zhu_jihong_ral} rather than a thin and deformable surgical suture.

Regarding these challenges, we propose a learning-driven framework for automated suture grasping task based on vision feedback from a calibrated stereo camera system, which is as shown in Fig.\ref{figure_intriduction_illustration}. Our main contributions are:
\textbf{1).} To improve the segmentation performance in surgical environments, we combine images from da Vinci robotic surgery with synthetic sutures, and build up our data set to train an offline model, which learns the general information of a surgical suture. 
Using a few online images, our model can quickly adapt to the present condition and achieve a higher performance for a segmentation task;
\textbf{2).} To compute 3D coordinates of a suture with arbitrary orientations, we developed a novel cost function and termination policies based on our previous method in \cite{lu_bo_tmech}\cite{lu_bo_2019_tase}. Using this upgraded approach, the ordering sequence of a suture (e.g. with self intersections or external crossings) can be completely obtained two camera frames;
\textbf{3).} We further construct a dense 3D vertex graph by calculating 3D positions of all pixel pairs along the suture, and optimize its spatial shape using an optimized shortest path approach. Consequently, the desired grasping point can be located based on the suturing criteria \cite{E_K_Batra}, proceeding an automated grasping operation.

The rest of this article is organized as follows: Section. \ref{section_methods} introduces the learning-driven workflow for a suture's grasping point computation. 
Section. \ref{section_experiments} presents the experimental validations concerning 2D segmentation and 3D grasping results. In Section. \ref{section_conclusion}, we will conclude the entire paper.

\begin{figure*}[t]
\centering
\includegraphics[width=0.95\textwidth]{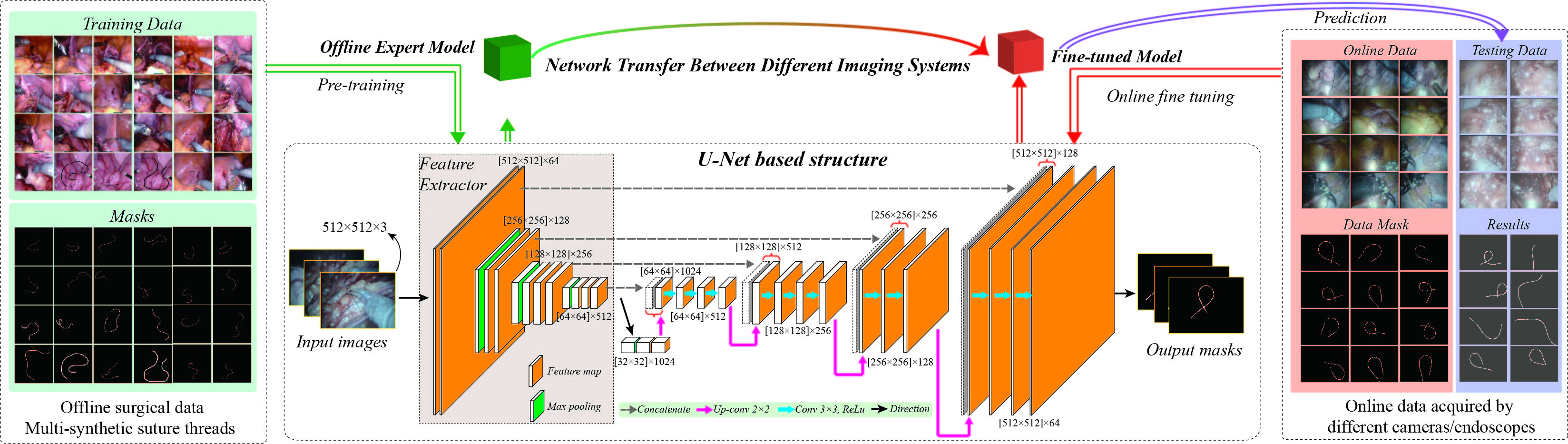}
	\caption{The training workflow for a suture segmentation in surgical environments. An offline data set containing legacy surgical data and synthetic sutures can be established, and it can be used to pre-train an expert model \(f_s\) which contains the general information of a suture. It can be expediently transferred to a new online scenario using a small amount of target domain data.}
	\label{figure_model_transfer_suture_segmentation}
\vspace{-0.5cm}
\end{figure*}

\section{Methodology}
\label{section_methods}
After cutting a suture during the interrupted suturing manipulation, the thread may arbitrarily drop above the surrounding tissue surface due to its nature of flexibility, and its precise 3D coordinates computation is essential for the following vision-based grasping task. To achieve this goal, we decomposed the overall procedure into 3 sub-tasks: \\
\textbf{1):} Suture segmentation from complex environments;\\
\textbf{2):} Ordering sequence analysis in stereo images;\\
\textbf{3):} 3D shape computation with optimization, and grasping point generation.

\subsection{Transferred Model for Online Suture Segmentation}
To segment a suture under different conditions, traditional imaging processing methods can hardly obtain satisfactory results under variant image scales, background noises, and ambient illuminations.
To achieve a higher performance of segmentation to guide the grasping task, a deep learning based model using transferred knowledge is proposed. 

The dominant objective of this model is to highlight a suture's curvilinear information, while removing surrounding noises.
U-Net is reported as a fast and precise model for image segmentation task, and it shows outstanding outcomes of intricate neuronal structure segmentation in electron microscopic stacks \cite{O_Ronneberger}. 
Here, U-Net model is migrated to our workflow as the backbone. The main training structure is shown in Fig. \ref{figure_model_transfer_suture_segmentation}. It is worth noticing that the model's first 11 layers are extended to 13 layers, constructing our feature extractor. By this modification, we can explicitly extract features within the image. We utilize the cross entropy as the loss function and froze the learned weights in the feature extractor to enhance the entire training efficiency.

\textit{1) Data Generation and Offline Training:}
To put against surgery noises, a large number of endoscopic images acquired from da Vinci robotic surgery are collected. 
Since limited images are obtained during surgical suturing, and to make the most of legacy surgical data, synthetic sutures are generated in these images to simulate suturing scenarios.

To ensure the diversity, sutures with different colors, scales and shapes are randomly generated, getting offline synthetic images \(\textbf{I}_{\textbf{OFF}}\). At the meanwhile, corresponding masks \(^{\textbf{M}}\textbf{I}_{\textbf{OFF}}\) can be assessed, and we can establish our expert database concerning surgical suture.
As shown in the left part of Fig. \ref{figure_model_transfer_suture_segmentation}, such information is imported to the training workflow, and a model \(f_s\), which learns the general information of a suture for surgical suturing, can be obtained.

\textit{2) Data Transfer and Online Prediction:} 
Barely using the expert model \(f_s\) for online segmentation can merely obtain a satisfactory result due to equipment and scenario differences. 
Even for dVRK systems, endoscopes may output images with variant qualities due to discrepancies caused by customized settings and long-time wear.
Thus, adapting to variances between different systems is a challenging task for our pre-trained model.

To address this problem, we employ a transfer learning method in our workflow. 
To build a refined model \(\widetilde{f}_s\), we transfer the learned weights in \(f_s\), then we froze the previous 13 layers in our feature extractor and fine-tune the remaining layers.
By using limited online annotated samples \(\textbf{I}_{\textbf{ON}}\) and \(^{\textbf{M}}\textbf{I}_{\textbf{ON}}\) from an on-site equipment, \(\widetilde{f}_s\) can be efficiently achieved for an online image prediction, which is illustrated in Fig. \ref{figure_model_transfer_suture_segmentation}.
It is defined that the online image is denoted as \(\textbf{I}_{\textbf{V}}\) and its annotated mask is \(^{\textbf{M}}\textbf{I}_{\textbf{V}}\), the fine-tuned model \(\widetilde{f}_s\) can obtain premium segmentation results for stereo cameras, and the results are expressed by \(\widetilde{\textbf{Se}}_l, \widetilde{\textbf{Se}}_r \in \mathbb{R}^2\). Validations with assessment metrics will be investigated in Section. \ref{section_experiments_segmentation_reconstruction}.

\subsection{Inference of Suture's Ordering Sequence}
\label{Section_ordering_inference}
Using \(\widetilde{\textbf{Se}}_l\) and \(\widetilde{\textbf{Se}}_r\), all contours in images can be picked out \(\{ \textbf{S}_1^l, \textbf{S}_2^l, ..., \textbf{S}_i^l,... \}\) \(\{ \textbf{S}_1^r, \textbf{S}_2^r, ..., \textbf{S}_j^r,... \}\). With manual indication in the left frame, we can implement the method in our previous paper \cite{lu_bo_tmech} to locate the cutting tip and obtain edge elements of the desired suture. Afterwards, these elements can be moved to the shape's centreline, which can be denoted as \(\hat{\textbf{S}}_L\) and \(\hat{\textbf{S}}_R\).

To accurately align the stereo pairs, the inference of a pixel-level sequence of a suture under multi-topologies is particularly important, since a small change in suture may result in different topologies as shown in Fig. \ref{figure_ordering_sequence_computation}(a).
Towards this goal, we designed a strategy which searches from the cutting tip and can be terminated automatically when the whole suture is traversed. 
For a 2D point \(V^t \in \hat{\textbf{S}}\) in the \(t^{th}\)searching loop\footnote{For a concise expression, we will omit the right subscript and only take the left frame for example in this section. In each searching loop, we can find a connecting point \(V^t\).}, we categorize all elements in \(\hat{\textbf{S}}\) as \textit{Detected Nodes} \(\mathbb{S}^t\), \textit{Active Nodes} \(\mathbb{A}^t\), and \textit{Far Nodes} \(\mathbb{F}^t\).
The connecting pixel is elected from \(\mathbb{A}^t\), which is defined as:
\begin{equation}
\begin{aligned}
    &\mathbb{A}^t = \{A_1^t, A_2^t, ..., A_j^t, ...\} \in \mathbb{R}^2 \\
    &s.t. \ \forall A_j^t \in \mathbb{A}^t: |A_j^t - V^t| \leq \Upsilon \ \& \ A_j^t \in \widetilde{\textbf{Se}}, A_j^t \notin \mathbb{S}^t
\end{aligned}
\end{equation}
where \(\Upsilon\) denotes the radius of the Region of Interest (RoI), \(A_j^t\) is one active node. 
Principles of the searching strategy are illustrated in Fig. \ref{figure_ordering_sequence_computation}(b)(c) and described as follows:
\begin{figure}[t]
\centering
\includegraphics[width=0.48\textwidth]{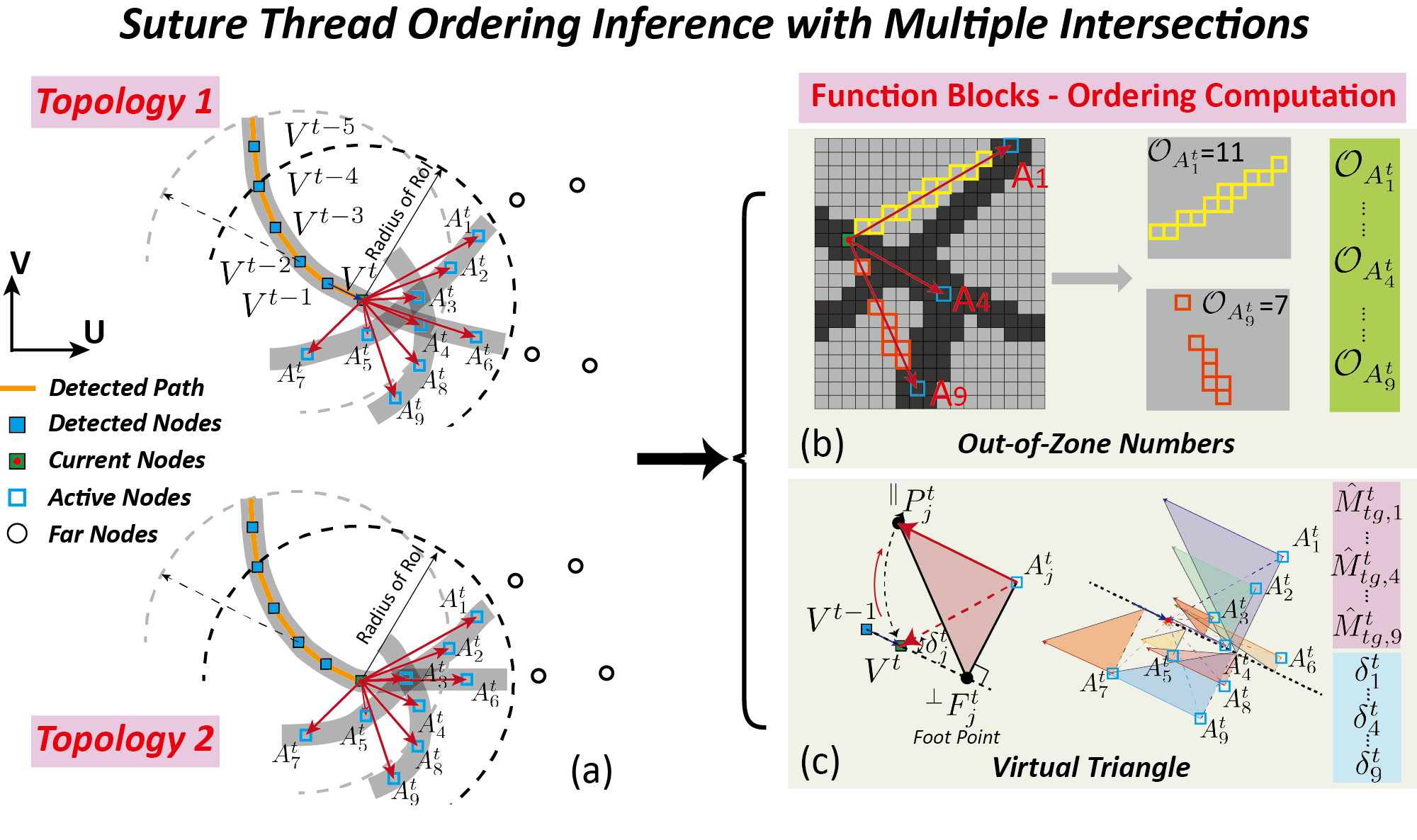}
	\caption{The cost function of the searching strategy for the ordering sequence inference of a suture with different shapes and intersections. Using our algorithm, the connecting pixels are \(A_4^t\) and \(A_3^t\) accordingly in two topologies which have slight differences between each other.}
	\label{figure_ordering_sequence_computation}
\vspace{-0.5cm}
\end{figure}

\textbf{{Principle 1:}}
We generate a line \(\textbf{l}_j^t\) by directly linking the candidate \(A_j^t\) and current pixel \(V^t\) as shown in Fig. \ref{figure_ordering_sequence_computation}.
\textbf{\textit{Definition:}} For any pixel \(\textsf{p}_{u_i,v_i} \in \textbf{l}_j^t\), if \(\textsf{p}_{u_i,v_i} \notin \widetilde{\textbf{Se}}\), \(\textsf{p}_{u_i,v_i}\) is an out-of zone element. The total number of such pixels along \(\textbf{l}_j^t\) is the \textit{\textbf{Out-of-Zone Numbers}} of candidate \(A_j^t\), which is denoted as \(\mathcal{O}_{A_j^t}\). For a potential candidate, its cost function regarding this term is designed following two criteria:\\
\textbf{\textit{Criterion 1:}} The cost value should be dramatically improved according to the increase of \(\mathcal{O}_{A_j^t}\) when it appearances.\\
\textbf{\textit{Criterion 2:}} 
If \(\mathcal{O}_{A_j^t}\) exceeds a particular value, we will treat the candidate \(A_j^t\) as a false connection, thereby the cost function should be insensitive to the further increase of \(\mathcal{O}_{A_j^t}\).

Taking both criteria into account, a logarithmic function is implemented to describe this principle. 

\textbf{{Principle 2:}}
For each candidate \(A_j^t\), we prefer the one has shorter distance to the current point \(V^t\). With the variation of \(|A_j^t - V^t |\) , the cost value changes following a linear manner.

\textbf{{Principle 3:}} 
For a continuous and smooth curvilinear object, the changing angle \(\delta_j^t\) between two adjacent vectors forming by three successive vertices should be evaluated. 
For each \(A_j^t\), we prefer the candidate generating a smaller value of \(\delta_j^t\), which ensures the smoothness along the whole shape. 
Therefore, we adopt an exponential function by putting higher cost when there exists shape change, while setting it insensitive among the low value domain.

\textbf{{Implement:}} 
To characterize \textbf{Principle 2} and \textbf{3}, a virtual triangulate formed by \(V^t\) and each candidate is adopted. As shown in Fig. \ref{figure_ordering_sequence_computation}(c), \(V^{t-1}\) and \(V^t\) are two successive vertices. For candidate \(A_j^t\), we have: point \(^{\perp}{F}_j^t \in {V^{t-1}V^t}\), \({^{\perp}{F}_j^i V^t} \perp {^{\perp}{F}_j^i A_j^t} \). Connecting \(A_j^t\) and \(V^t\), we obtain line \(\overrightarrow{A_j^t V^t}\) and variation angle \(\delta_j^t\). By rotating this vector within acute angle until reaching the point \(^{\parallel} P_j^t\), which satisfies \( \overrightarrow{ {^{\parallel}P_j^t}{A_j^t} } \parallel \overrightarrow{^{\perp}{F}_j^i V^t}\), a virtual triangle \(\triangle {^{\perp}{F}_j^i}{^{\parallel}P_j^t}A_j^t\) can be generated, and its virtual area can be implemented as the partial term among the whole function, which is:
\begin{equation}
\begin{aligned}
	&M_{tg, j}^t = 0.5\cdot \underbrace{|A_j^t - V^t|}_{Side A} \cdot \underbrace{{(|A_j^t - V^t| \cdot\sin(\delta_j^t)}}_{Side B}\\
	& \textit{s.t.} \ \delta_j^t = \arccos{\frac{(A_j^t - V^t)\cdot (V^t - V^{t-1})}{|A_j^t - V^t|\cdot|V^t - V^{t-1}|}}, t>1 \subseteq \mathbb{Z}
\end{aligned}
\end{equation}

To represent a monotonically increasing property with respect to \(\delta_j^t \in [0,\pi]\), the term \(\sin(\delta_j^t)\) in virtual \textit{Side B} controlling the changing direction can be optimized as: \(\sin(\delta_j^t/2)\), and we further adopt an exponential function as claimed in \textbf{Principle 3} to increase costs of sharp variations. Taking these factors into account, the optimized virtual area \(\hat{M}_{tg,j}^t\) can be re-designed as:
\begin{equation}
	\hat{M}_{tg,j}^t = 0.5\cdot {|A_j^t - V^t|} \cdot e^{\sin(\delta_j^t / 2) }
\end{equation}

With the term of \textbf{Principle 1}, the composite cost function for each candidate at \(t^{th}\) searching loop can be derived as:
\begin{equation}
\begin{aligned}
	&\textsf{C}_j^t = \log(\mathcal{O}_{A_j^t} + \mu) + \hat{M}_{tg,j}^t \xrightarrow{Tune} \textsf{C}_j^t(A_j^t, \epsilon_1, \epsilon_2, \epsilon_3) \\
	& =\log(\epsilon_1 \cdot \mathcal{O}_{A_j^t} + \mu) + \epsilon_2 \cdot {|A_j^t - V^t|} \cdot e^{\epsilon_3 \cdot \sin(\frac{\delta_j^t}{2}) }
\end{aligned}
\end{equation}
where \(\mu\) is a small positive gain, and \(\{\epsilon_1, \epsilon_2, \epsilon_3\}\) are tuning parameters that determines the influences of three principles in the function. During each searching loop, the connecting pixel can be achieved by:
\begin{equation}
\label{eq_seraching_policy}
	\forall A_j^t \in \mathbb{A}^t: \ V^{t+1} = \mathop{\arg\min}_{A_j^t} \{\textsf{C}_j^t(A_j, \epsilon_1, \epsilon_2, \epsilon_3) \}
\end{equation}
The inferred ordering sequence of a suture can be thereby obtained as \(\mathbb{S} = \{ V^1, V^2, ..., V^k, ... \}\). 
To enable a robust scheme for different cases, three tuning parameters can be autonomously adjusted within their given ranges to maximize the element number in \(\mathbb{S}\). Besides, termination policies are also proposed, and taking Eq. (\ref{eq_seraching_policy}) into account, tuning parameters can be selected as:
\begin{equation}
\label{eq_tunning_and_termination}
\small
\begin{aligned}
&\forall \{ \epsilon_1 \in \textbf{\textsf{R}}_1, \epsilon_2 \in \textbf{\textsf{R}}_2, \epsilon_3 \in \textbf{\textsf{R}}_3 \} : \epsilon_1, \epsilon_2, \epsilon_3 = \mathop{\arg\max} \{|\mathbb{S}| \}\\
&s.t. \ \mathbb{A}^t \neq \varnothing : \mathcal{O}_{A_j^t} < \tau_O \ \& \ {\delta}_j^t < \tau_V
\end{aligned}
\end{equation}
where \(\textbf{\textsf{R}}_1 \in [1,5], \textbf{\textsf{R}}_2 \in [0.1, 0.5], \textbf{\textsf{R}}_3 \in [0.02, 0.1]\) denote values' ranges, and \(\tau_O = 2.6\Upsilon, \tau_V = 2\pi/3\) are the terminating thresholds in our research.
Following the parallel optical axis assumption, we further apply this algorithm to both frames to get stereo pairs \([\mathbb{S}_l \leftrightarrow \mathbb{S}_r]\) and 3D dense points, i.e. vertices \(\textbf{V}^{1 \mapsto \omega_1} = \{ \textbf{V}^1, \textbf{V}^2, ..., \textbf{V}^{\omega_1} \} \in \mathbb{R}^3 \) along the suture using the method in our previous article \cite{lu_bo_2019_tase}.

\subsection{Online Optimization of 3D Shape}
Due to limited resolution of camera/endoscope, there may exist zigzag routes by connecting the computed dense 3D points one by one, thus forming undesired configurations. Besides, we also need to compensate for unexpected errors resulting from noises in set \(\textbf{V}^{1 \mapsto \omega_1}\) in order to obtain precise 3D representative vertices \(\hat{\textbf{V}}^{1 \mapsto \omega_2} \subseteq \textbf{V}^{1 \mapsto \omega_1}\) of the suture.

\begin{figure}[h]
\centering
\includegraphics[width=0.48\textwidth]{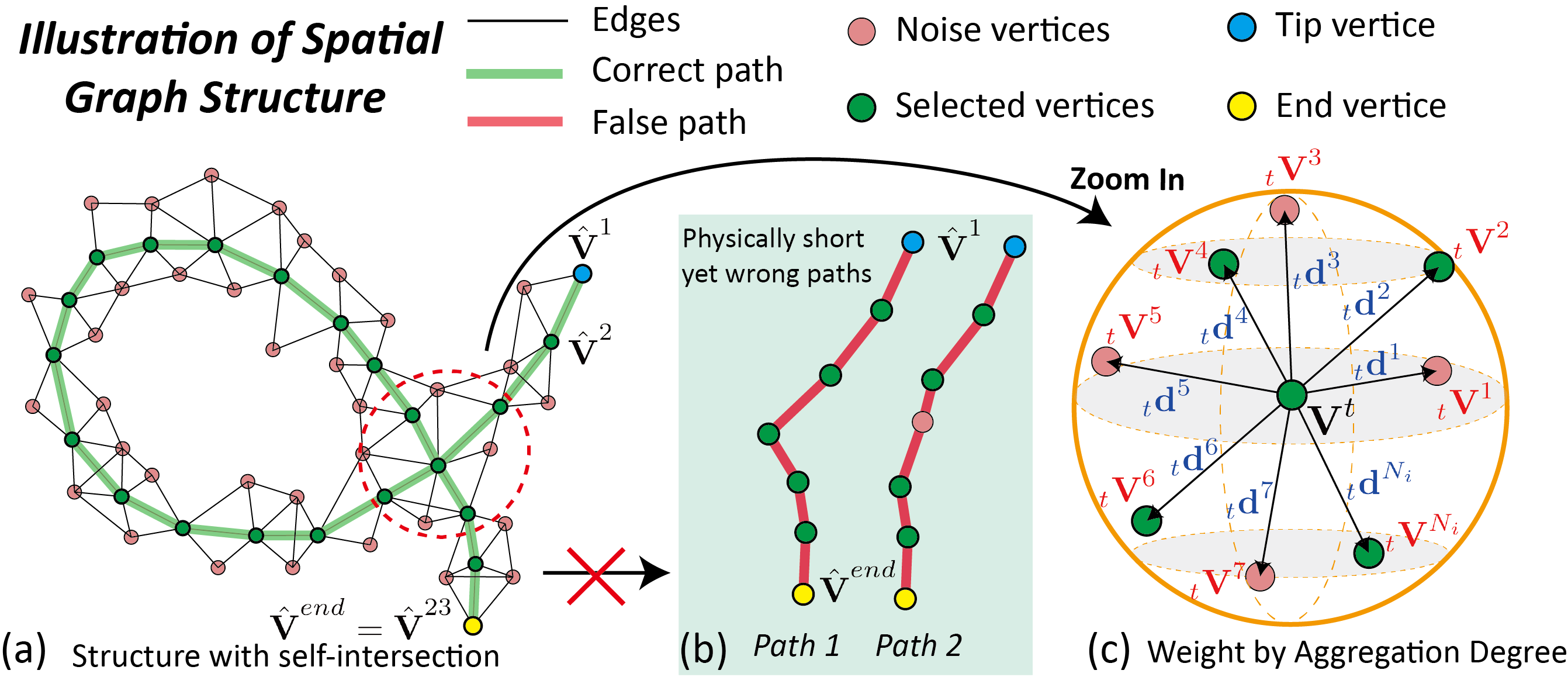}
	\caption{Spatial shape optimization of a suture with self-intersections based on a shortest path method with optimized and composite edges' weights.}
	\label{figure_3D_optimisation}
\end{figure}

To realize this objective, the approach based on Dijkstra's shortest path theory \cite{N_Jasika} is developed to optimize the 3D shape between the tip and end pints. 
For a self-intersected suture as shown in Fig. \ref{figure_3D_optimisation}(a) and (b), the traditional method gives a result of minimal physical length between two ends, which turns to the distal vertice at the crossing space. However, such results indicate a fragmentary outcome of the suture, which may jeopardize the grasping point computation.

To characterize the overall information of suture, we first construct the 3D edge structure using all vertices in \(\textbf{V}^{1\rightarrow \omega_1}\), and remove those edges whose lengths are larger than \(\tau_L\):
\begin{equation}
\label{eq_physical_graph_construction}
\small
	\forall \{\textbf{V}^i,\textbf{V}^t \}\in \textbf{V}^{1\rightarrow\omega_1}: \textsf{D}^{i \leftrightarrow t} =
\left\lbrace
\begin{aligned}
|\textbf{V}^i &- \textbf{V}^t|, &|\textbf{V}^i - \textbf{V}^t| \leqq \tau_L\\
&\infty , &|\textbf{V}^i - \textbf{V}^t| > \tau_L
\end{aligned}
\right.
\end{equation}
where \(\textsf{D}^{i \leftrightarrow t}\) denote physical length of Edge \(\textsf{E}^{i \leftrightarrow t}\) between vertice \(\textbf{V}^i\) and \(\textbf{V}^t\). By treating diagonal elements as \(\infty\), the complete 3D length matrix \(\textbf{\textsc{D}}_{\textbf{\textsf{G}}}\) can be formulated as:
\begin{equation}
\centering
\textbf{\textsf{D}}_{\textbf{\textsf{G}}} = 
\left[
\begin{aligned}
\centering
&\infty, {\textsf{D}}^{1\leftrightarrow 2}, {\textsf{D}}^{1\leftrightarrow 3}, ..., {\textsf{D}}^{1\leftrightarrow \omega_1}\\
&{\textsf{D}}^{2\leftrightarrow 1}, \infty, {\textsf{D}}^{2\leftrightarrow 3}, ..., {\textsf{D}}^{2\leftrightarrow \omega_1}\\
&..., ..., {\textsf{D}}^{i\leftrightarrow i} = \infty, ..., {\textsf{D}}^{i\leftrightarrow \omega_1}\\
&{\textsf{D}}^{\omega_1 \leftrightarrow 1}, ..., ..., {\textsf{D}}^{\omega_1 \leftrightarrow \omega_1 -1}, \infty
\end{aligned}
\right]_{\omega_1 \times \omega_1}
\end{equation}

To further enhance the representative capability of the graph structure, we design a composite weight matrix \(\textbf{\textsf{W}}_{\textbf{\textsf{G}}}\), which denotes the penalty for each \(\textsf{E}^{i \leftrightarrow t}\) by considering the vertice density around it two ends \(\textbf{V}^i\) and \(\textbf{V}^t\), as well as evaluating their sequential difference.

We find out the valid vertice \(_t\textbf{V}^j\) within a virtual sphere, which is centered by the point \(\textbf{V}^t\) with a radius \(\Upsilon_3\) as shown in Fig. \ref{figure_3D_optimisation}(c).
Counting the total number of \(_t\textbf{V}^j\) within this space, we can formulate a penalty following the right part of the Gaussian curve, which determines the clustering level of computed points around \(_t\textbf{V}^j\).
Apart from this, we measure all distances between surround points to \(\textbf{V}^t\). The total penalty for vertice \(\textbf{V}^t\) can be designed as:
\begin{equation}
\label{eq_vertice_p1_penalty}
\small
\begin{aligned}
\textsf{P}^t_1&=\underbrace{\frac{1}{\sqrt{2\pi\ \sigma_1}}\cdot \exp{(-\frac{{N}_i}{2 \sigma_1})}}_{Effect \ of \ Number}\cdot \underbrace{\sum_{j}^{{N}_{\textbf{i}}}\frac{S_f}{{N}_{\textbf{i}}}(\log(|_t\textbf{V}^j - \textbf{V}^t| + \nu_1)}_{Effect \ of \ Distance}\\
& = \frac{S_f}{N_i} \cdot \exp{(-\frac{{N}_i}{2 \sigma_1})} \cdot \sum_{j}^{{N}_{\textbf{i}}}(\log(|_t\textbf{V}^j - \textbf{V}^t| + \nu_1)
\end{aligned}
\end{equation}
where \(S_f\) denotes a scaling factor, \(\sigma_1\) determines the steepness of the Gaussian curve, and \(\nu_1\) is a positive constant. Multiplying these values of two points \(\textbf{V}^i\) and \(\textbf{V}^j\), the first term of the penalty for Edge \(\textsf{E}^{i \leftrightarrow t}\) forming by these two vertices can be obtained as:
\begin{equation}
\label{eq_edge_penalty_1}
	{\textsf{P}}^{i\leftrightarrow t}_1 = \textsf{P}^t_1 \cdot \textsf{P}^i_1
\end{equation}

Based on the results in Section \ref{Section_ordering_inference}, the ordering differences between two ends vertices of Edge \(\textsf{E}^{i \leftrightarrow t}\) should be evaluated, which aims to avoid unexpected turns around crossing nodes and get a complete suture's shape. Thus, the second term of the penalty for Edge \(\textsf{E}^{i \leftrightarrow t}\) is designed as:
\begin{equation}
\label{eq_edge_penalty_2}
\begin{aligned}
\textsf{\textsf{P}}^{i\leftrightarrow t}_2 = 1 + \exp(\frac{|i-t|}{20} - 2)
\end{aligned}
\end{equation}

Combining \(\textsf{P}^{i \leftrightarrow t}_1\) and \(\textsf{P}^{i \leftrightarrow t}_2\), the composite weight for Edge \(\textsf{E}^{i \leftrightarrow t}\) in the graph structure can be expressed by:
\begin{equation}
\begin{aligned}
\forall \textbf{V}^i, \textbf{V}^t \in \textbf{V}^{1 \leftrightarrow \omega_1}: {\textsf{W}}^{i\leftrightarrow t} = {\textsf{P}}^{i\leftrightarrow t}_1 \cdot {\textsf{P}}^{i\leftrightarrow t}_2
\end{aligned}
\end{equation}

The overall weight matrix \(\textbf{\textsf{W}}_{\textbf{\textsf{G}}}\) is hence formulated as:
\begin{equation}
\textbf{\textsf{W}}_{\textbf{\textsf{G}}} = 
\left[
\begin{aligned}
&1, {\textsf{W}}^{1\leftrightarrow 2}, {\textsf{W}}^{1\leftrightarrow 3}, ..., {\textsf{W}}^{1\leftrightarrow \omega_1}\\
&{\textsf{W}}^{2\leftrightarrow 1}, 1, {\textsf{W}}^{2\leftrightarrow 3}, ..., {\textsf{W}}^{2\leftrightarrow \omega_1}\\
&..., ..., {\textsf{W}}^{j\leftrightarrow j} = 1, ..., ...\\
&{\textsf{W}}^{\omega_1 \leftrightarrow 1}, ..., ..., {\textsf{W}}^{\omega_1 \leftrightarrow \omega_1 -1}, 1
\end{aligned}
\right]_{\omega_1 \times \omega_1}
\end{equation}

We integrate \(\textbf{\textsf{D}}_{\textbf{\textsf{G}}}\) and \(\textbf{\textsf{W}}_{\textbf{\textsf{G}}}\), and the optimized graph matrix can be attained using the Hadamard product, which is:
\begin{equation}
	\textbf{\textsf{G}}(\textsf{E}, \textbf{V}) = \textbf{\textsf{D}}_{\textbf{\textsf{G}}} \odot \textbf{\textsf{W}}_{\textbf{\textsf{G}}}
\end{equation}

Putting \(\textbf{\textsf{G}}(\textsf{E}, \textbf{V})\) into Dijkstra's shortest path method, the shortest path containing the updated vertices \(\hat{\textbf{V}}^{1 \leftrightarrow \omega_2} = \{\hat{\textbf{V}}^1, \hat{\textbf{V}}^2, ..., \hat{\textbf{V}}^{\omega_2}\} \) can be obtained.
The pseudocode of this optimization method is listed in Algorithm. \ref{algorithm_3D_shape_optimization}.
\begin{algorithm}
\label{algorithm_3D_shape_optimization}
\caption{Spatial shape optimization of a suture}
\KwData{3D vertices \(\textbf{V}^{1 \leftrightarrow \omega_1}\), the radius \(\Upsilon_3\) of a virtual sphere, parameter \(\tau_L\) in Eq. (\ref{eq_physical_graph_construction}).}
\For{\(\textbf{V}^t \in \textbf{V}^{1 \leftrightarrow \omega_1} \stackrel{find}{\longrightarrow}\)}{
\(\forall\textbf{V}^i \in \textbf{V}^{1 \leftrightarrow \omega_1} \stackrel{find}{\longrightarrow}: |\textbf{V}^i -\textbf{V}^t \leqslant \Upsilon——3|\rightarrow\) Denoted as: \(\{ _t\textbf{V}^{1}, _t\textbf{V}^{2}, ..., _t\textbf{V}^{N_i} \}\)\;
Computed \(\{ _t\textbf{d}^{1}, _t\textbf{d}^{2}, ..., _t\textbf{d}^{N_i} \}\) w.r.t \(\textbf{V}^t\) (Fig. \ref{figure_3D_optimisation})\;
Computed \(\textsf{P}_1^t \leftarrow\) Eq. (\ref{eq_vertice_p1_penalty})\;
}
\(\forall \{\textbf{V}^t,\textbf{V}^i\} \in \textbf{V}^{1 \leftrightarrow \omega_1}, t \neq i \longrightarrow \) Compute their edge penalty \({\textsf{P}}^{i\leftrightarrow t}_1, {\textsf{P}}^{i\leftrightarrow t}_2\) using Eq. (\ref{eq_edge_penalty_1}) (\ref{eq_edge_penalty_2})\;
Calculate \({\textsf{W}}^{i\leftrightarrow t}\longrightarrow\) Form weight matrix \(\textbf{\textsf{W}}_{\textbf{\textsf{G}}}\)\;
\While{(True)}{
Generate distance matrix \((\textbf{\textsf{D}}_{\textbf{\textsf{G}}})_{\omega_1 \times \omega_1}\)\;
\(\forall \textsf{D}^{i \leftrightarrow t} \in \textbf{\textsf{D}}_{\textbf{\textsf{G}}} \longrightarrow \textsf{D}^{i \leftrightarrow t} = \infty\), when \(|\textbf{V}^t - \textbf{V}^i| > \tau_L\) or \(i=t\)\;
\(\textbf{\textsf{G}}(\textsf{E}, \textbf{V}) = \textbf{\textsf{D}}_{\textbf{\textsf{G}}} \odot \textbf{\textsf{W}}_{\textbf{\textsf{G}}} \stackrel{Dijkstra}{\longrightarrow}\) Shortest path \(L_{min}\) and \(\hat{\textbf{V}}^{1 \leftrightarrow \omega_2} = \{\hat{\textbf{V}}^1, \hat{\textbf{V}}^2, ..., \hat{\textbf{V}}^{\omega_2}\} \) of the suture\;
\If{\(L_{min} \neq \infty\)}{
break\;
}
\(\tau_L =\tau_L+0.5 \leftarrow\) \footnotesize{\textsf{In case there is no path between two ends, we construct more connecting edges in the graph.}}
}
\KwResult{Updated vertices \(\hat{\textbf{V}}^{1 \leftrightarrow \omega_2} = \{\hat{\textbf{V}}^1, \hat{\textbf{V}}^2, ..., \hat{\textbf{V}}^{\omega_2}\} \) and minimal length \(L_{min}\) of the suture.}
\end{algorithm}

Given the entire 3D information of a suture, the grasping point \(\textsf{GS}_{\textsf{C}}=[\textsf{GX}_{\textsf{C}}, \textsf{GY}_{\textsf{C}}, \textsf{GZ}_{\textsf{C}}, 1]^{\top} \) with respect to the camera coordinate can be calculated by reserving a sufficient length for the looping manipulation, thereby proceeding the automated grasping with the Euclidean target in the task space produced by \(\textsf{GS}_{\textsf{R}} = {{_{\textsf{C}}}\textbf{\textsf{T}}}^{\textsf{R}}\cdot \textsf{GS}_{\textsf{C}} \), in which \({{_{\textsf{C}}}\textbf{\textsf{T}}}^{\textsf{R}} \in \textbf{SE}(3)\) is the transformation matrix from camera coordinate to robot frame obtained by a standard hand to eye calibration approach \cite{C_Wengert}.

\section{Experimental Results}
\label{section_experiments}
In this section, we extensively validated the performance of the deep learning-based suture segmentation and the feasibility of the sequence inference algorithm. 
Based on the stereo visual perception, experiments related to the 3D suture reconstruction and automated grasping were simulated in V-REP using the dVRK model introduced by \cite{G_A_Fontanelli}. The overall framework were tested by a UR robot and dVRK system.

\subsection{Performance Assessment of Suture Segmentation}
\label{section_experiments_segmentation_reconstruction}
\begin{figure*}[h]
\centering
\includegraphics[width=0.95\textwidth]{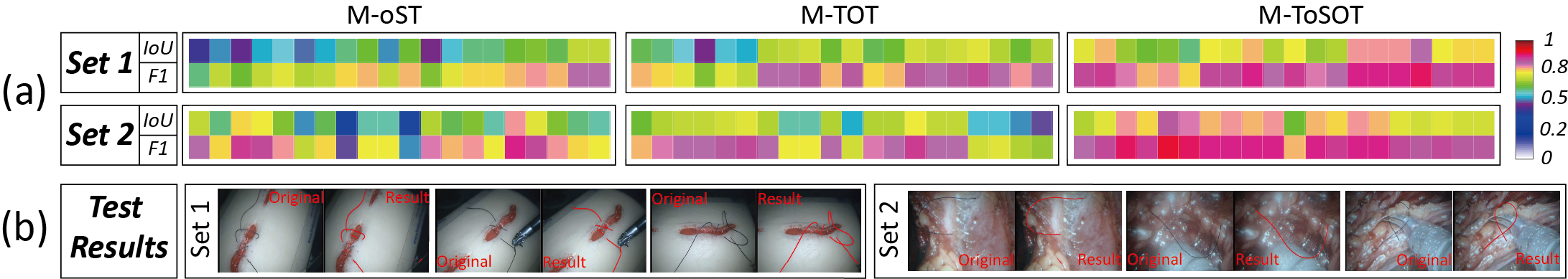}
	\caption{(a). Segmentation results regarding IoU and F1 scares of 3 different models based on 40 test images obtained by two dVRK systems; (b). Typical segmentation results. Surgical phantoms were adopted in Set 1, and color-printed surgical scenes were used in Set 2 to simulated complex scenarios.}
	\label{figure_qualitative_segmentation_results}
\vspace{-0.5cm}
\end{figure*}
Our learning model was constructed based on Keras framework with a Tensorflow backbend, and it was run on a Nvidia Titan 1080 GPU. 
Using the U-Net based structure with a transfer learning strategy, we investigated and compared the suture segmentation performance of three models, which are: Models from (1). offline synthetic images training (\textit{\textsf{M-oST}}); (2). training with online target domain images (\textit{\textsf{M-TOT}}); (3). transfer learning from offline synthetic data to online target domain ((\textit{\textsf{M-ToSOT}})). The metrics for the segmentation performance assessment are Intersection over Union (IoU), Precision, Recall, and F1 Score, which are defined as:
\begin{equation}
\small
IoU = \frac{|\widetilde{\textbf{Se}} \bigcap ^{\textbf{M}}\textbf{I}_{\textbf{V}}|}{|\widetilde{\textbf{Se}}| + |^{\textbf{M}}\textbf{I}_{\textbf{V}}| - |\widetilde{\textbf{Se}} \bigcap ^{\textbf{M}}\textbf{I}_{\textbf{V}}|}, F1 = \frac{ 2\cdot Pre \cdot Rec}{Pre + Rec}
\end{equation}
where \(Pre\) and \(Rec\) denote the segmentation precision and recall, which are accordingly expressed as \(|\widetilde{\textbf{Se}} \bigcap ^{\textbf{M}}\textbf{I}_{\textbf{V}}|/|\widetilde{\textbf{Se}}| \) and \(|\widetilde{\textbf{Se}} \bigcap ^{\textbf{M}}\textbf{I}_{\textbf{V}}|/|^{\textbf{M}}\textbf{I}_{\textbf{V}}| \).

We built up the offline model using 200 groups of data (the number can be further extended to increase the model's generality), and 40 target domain images were used to fine-tune the model. Experiments were tested on 40 images captured from 2 different dVRK endoscopic camera manipulators. To add variances to images, the target domain was a phantom-based suturing scenario in Set 1, and surgical scenes were color printed and treated as backgrounds in Set 2.
The quantitative results are listed in Table. \ref{table_quantitative_results_segmentation}.
As noticed, \textit{\textsf{M-sOT}} is unstable in the target domain testing. Its high precision with small IoU and recall tell that it only segments scattered pieces in the image. By training directly with limited online data from target domain, \textit{\textsf{M-TOT}} is inferior to \textit{\textsf{M-ToSOT}} with respects to all assessment metrics. 
\begin{table}[h]
\caption{Comparisons of the learning based suture segmentation performance using three models.}
\label{table_quantitative_results_segmentation}
\begin{center}
\begin{tabular}{|c |c | c|| c| c| c | }
\hline
\multicolumn{3}{|c||}{{\textsf{\textbf{Model}}}} 
&\textit{\textsf{M-oST}} 
&\textit{\textsf{M-TOT}} 
&\textit{\textsf{M-ToSOT}}\\
\hline
\hline

\multirow{3}{*}{\textit{\textsf{Set 1}}} &\multirow{1}{*}{\textit{\textsf{IoU}}} & \textsf{ave} &55.8\% &62.9\% &\textbf{71.1\%} \\
\cline{2-6}

&\multirow{1}{*}{\textit{\textsf{Pre}}} & \textsf{ave} &\textbf{84.6}\% &63.2\% &72.5\%  \\
\cline{2-6}

&\multirow{1}{*}{\textit{\textsf{Rec}}} & \textsf{ave} &61.8\% &96.2\% &\textbf{97.3\%}  \\
\cline{2-6}

\hline
\hline

\multirow{3}{*}{\textit{\textsf{Set 2}}} &\multirow{1}{*}{\textit{\textsf{IoU}}} & \textsf{ave} &59.8\% &60.8\% &\textbf{72.1\%} \\
\cline{2-6}

&\multirow{1}{*}{\textit{\textsf{Pre}}} & \textsf{ave} &\textbf{79.9}\% &62.3\% &73.5\%  \\
\cline{2-6}

&\multirow{1}{*}{\textit{\textsf{Rec}}} & \textsf{ave} &71.0\% &96.3\% &\textbf{97.5\%}  \\
\cline{3-6}
\hline
\end{tabular}
\end{center}
\end{table}

In Fig. \ref{figure_qualitative_segmentation_results}(a), we listed qualitative results of IoU and F1 Scores for each individual test. Using limited data without prior knowledge, \textit{\textsf{M-TOT}} appears to be unstable across its overall test set.
Regarding the numbers in Table. \ref{table_quantitative_results_segmentation}, in which \textit{\textsf{M-ToSOT}} achieves the best performance in all aspects, especially for the highest recall number 97.5\%. Although it contains false positives in the tests, our following sequence inference algorithm can compensate for this shortcoming. Until now, the transfer learning strategy can be concluded workable for our suture segmentation task, and typical results of suture segmentation were shown in Fig. \ref{figure_qualitative_segmentation_results}(b).

\subsection{Validations of Vision-based Automated Grasping }
Based on segmentation results, we will validate the feasibility of the sequence inference algorithm of suture, and further examine the performance of this vision-guided automated grasping manipulation in this section.

\begin{figure}[t]
\centering
\includegraphics[width=0.47\textwidth]{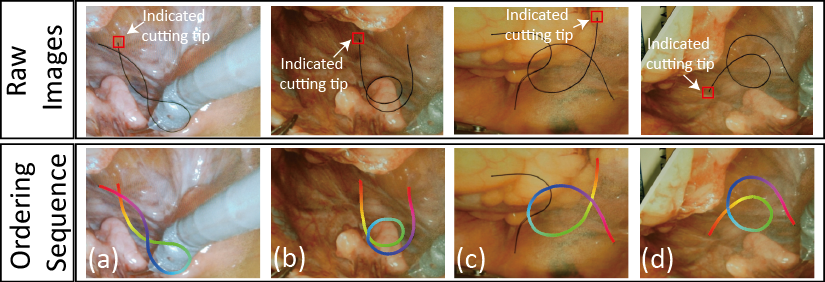}
	\caption{Typical results of the ordering sequence inference of a suture under multiple topologies and variant conditions.}
	\label{figure_suture_sequence_inference}
\vspace{-0.5cm}
\end{figure}

\textit{\textbf{1).} Ordering Sequence Inference:} To begin the test, we collected image data by putting sutures on color-printed surgical scenes and orientating them with multiple shapes. After sending the image to the fine-tuned model \textit{\textsf{M-ToSOT}}, all suture-like objects \(\widetilde{\textbf{Se}}\) will be segmented.
To simplify the testing procedure, we selected suture's one end as the cutting tip, and indicated a RoI around it as illustrated in Fig. \ref{figure_suture_sequence_inference}. 

Employing the method in our previous paper \cite{lu_bo_tmech}, the precise tip in a stereo-camera could be found, and the inference can be thereby proceeded. 
As also shown in Fig. \ref{figure_suture_sequence_inference}, four typical results concerning ordering inference were presented, and gradient colors represent the calculated sequence along the suture. It can be notice, the loss function based searching algorithm can efficiently figure out the correct direction with suture's self intersections. Even crossed by other sutures, which was the case in Fig. \ref{figure_suture_sequence_inference}(c), the shape can be correctly inferred and the computation can be terminated by Eq. (\ref{eq_tunning_and_termination}).

\textit{\textbf{2).} Key Points Approaching Along the Suture:}
\begin{figure*}[h]
\centering
\includegraphics[width=0.95\textwidth]{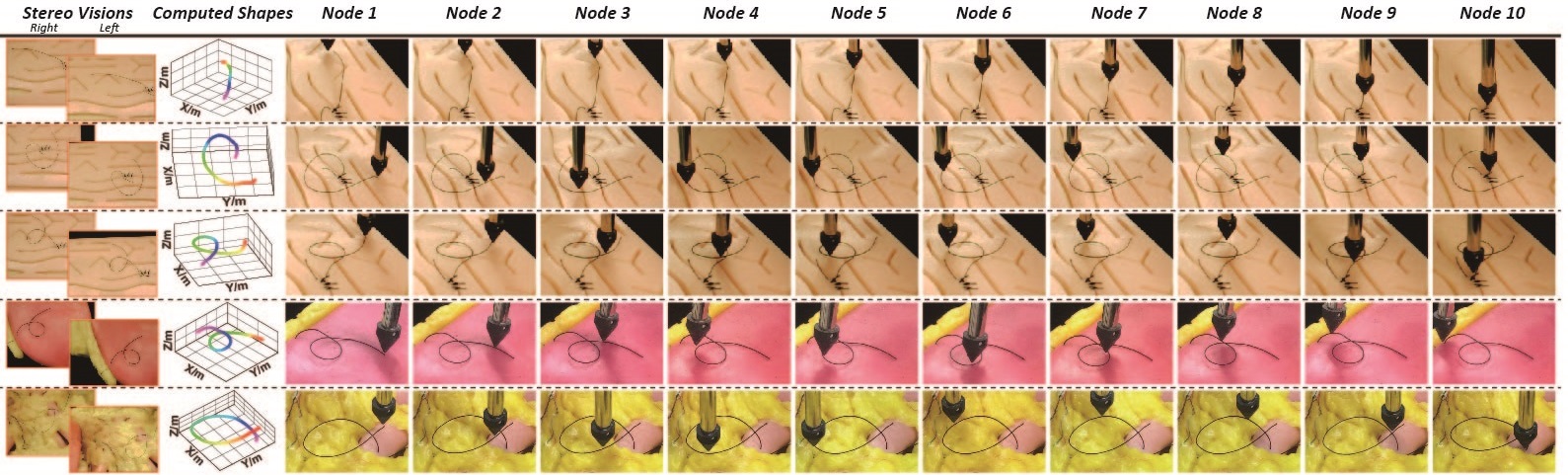}
	\caption{The 3D reconstruction of a suture and a vision-based "grasping". We attached an end effector to UR5 to simulate the grasping task by sending the instrument's tip to follow the selected key points along the suture. Different artificial tissues were adopted as background.}
	\label{figure_approaching_suture_UR5}
\end{figure*}
\begin{table*}[h]
\centering
\caption{Experimental results of suture "grasping" by approaching key nodes along the suture - Error amount, unit: \(mm\)}
\label{table_erros_of_approaching_suture_key_nodes}
\begin{tabular}{ | c || c c c c c c c c c c || c | c| }
\hline
\textsf{\textbf{Group}} &\multicolumn{10}{ |c||}{\textsf{\textbf{Selected Key Nodes 1 - 10}}} &\textsf{\textbf{Ave}} & \textsf{\textbf{Max}} \\
\hline
\hline
\rowcolor{lightgray} Straight Suture &3.2 & 0.6 &1.2 &1.9 &0.7 &1.8 &3.6 &3.5 &3 &2.6 &2.21 &3.6\\
\hline
Curved Suture   &0.8 & 0.3 &2.3 &1.9 &2.4 &1.6 &2.1 &0.8 &1.3 &0.8 &1.43 &2.4\\
\hline
\rowcolor{lightgray} Intersected Suture &1.8 &4.1 &1.4 &1.1 &0.8 &2 &1.5 &2.8 &1.2 &1.8 &1.85 &4.1\\
\hline
Artificial Liver as background &1.6 &1.4 &2.8 &2.7 &2.2 &1.7 &0.9 &0.5 &1.3 &1.1 &1.62 &2.8\\
\hline
\rowcolor{lightgray} Surgical Phantom as background &1.5 &1.6 &1.5 &1.2 &1.3 &3.1 &1.0 &2.6 &4.5 &1.3 &1.96 &4.5\\
\hline
\end{tabular}
\end{table*}
In this part, we built up a vision-based robotic system and extensively evaluated the computational accuracy along the whole suture. As shown in Fig. \ref{figure_approaching_suture_UR5}, sutures were stitched through different surgical phantoms, possessing various orientations. 
Using a calibrated stereo camera, we figured out its spatial coordinates in robot base using the information from an offline hand-to-eye calibration \cite{C_Wengert}. 
We attached a shaft with touching tip to the end of a UR5 robot, and the transformation matrix \({{_{\textsf{C}}}\textbf{\textsf{T}}}^{\textsf{R}}\) was reformulated by adding a translational vector in \(Z\) direction, while the orientation of the end effector was set vertical to the surface. 

Key points were evenly selected from calculated results, and the touching tip was repeatedly driven to these positions to simulate a grasping operation and validate computational results concerning the suture's spatial coordinates. Key frames in 5 experiments were shown in Fig. \ref{figure_approaching_suture_UR5}, and errors between the tip and the suture were recorded in Table. \ref{table_erros_of_approaching_suture_key_nodes}. As noticed that average errors in each group were all below \(2.3mm\), and the maximum one happened in the \(5th\) set, which was \(4.5mm\). Considering errors induced from calibrations, such accuracy can be tolerated with a \(10mm\)-long dVRK forcep when grasping a suture.

\textit{\textbf{3).} Feasibility Study by Simulation in V-REP and dVRK System:}
\begin{figure}[b]
\centering
\includegraphics[width=0.45\textwidth]{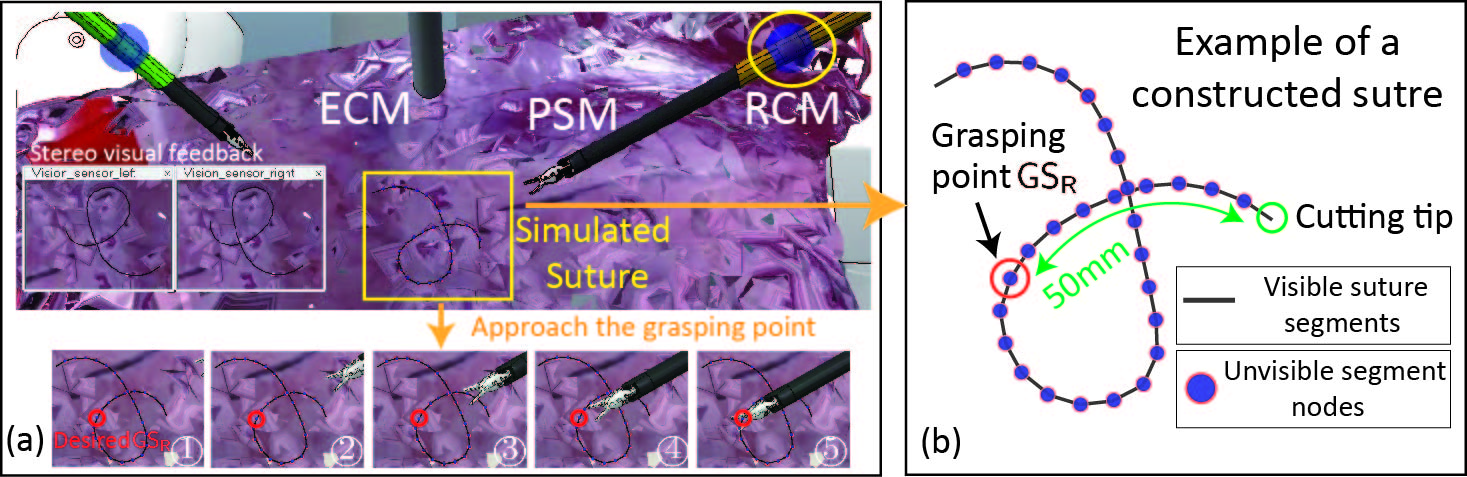}
	\caption{Simulated suture grasping in V-REP. (a). Illustrations of a simulation environment using point cloud meshes and one simulated grasping process based on a dVRK model; (b). Example of a constructed suture in V-REP.}
	\label{figure_simulated_grasping}
\end{figure}
To evaluate the entire framework, we implemented it to V-REP simulator for validation. 
In the simulator, we built up sutures with different shapes by connecting various pieces of cylinders, whose length is \(5mm\) with a radius of \(0.5mm\) on the cross section. To characterize a smooth and slender suture, every two pieces were connected by a spherical joint, which was set invisible in the stereo vision sensors.
Besides, 3D point clouds computed based on surgical videos were added to act as the background in space, and one typical case was shown in Fig. \ref{figure_simulated_grasping}(a).

In the experiment, the grasping point reserved a \(50mm\) length to the cutting tip as shown in Fig .\ref{figure_simulated_grasping}(b), thus obtaining the grasping node \(\textsf{GS}_{\textsf{C}}\) in the camera coordinate. With a dVRK model introduced in \cite{G_A_Fontanelli}, we can conveniently get the transformation relationship between PSM and ECM, and the grasping point in the robot space \(\textsf{GS}_{\textsf{R}}\) can be acquired. 

We conducted 8 experiments in 2 test sets by orientating a suture with different shapes and lengths. Based on the computational results, a moving trajectory using the linear interpolation can be generated. 
Only with stereo visions and system's calibrations, the errors between \(\textsf{GS}_{\textsf{R}}\) (3D point of the end effector) and the desired grasping point (obtained by reading the 3D coordinates of the correct node) were listed in Table. \ref{table_suture_approaching_and_simulated_grasping}. It is worth noticing that the average errors in two testing sets were only \(2.33mm\) and \(1.73mm\), and even the largest error was below \(3.5mm\), which is quite precise for a dVRK grasper to pick up the suture. 
\begin{table}[t]
\caption{Experimental results of suture "grasping" by approaching key nodes along the suture - Error amount, unit: \(mm\)}
\label{table_suture_approaching_and_simulated_grasping}
\centering
\begin{tabular}{ | c || c c c c c c c c | }
\hline
\textsf{\textbf{Test Topology}} &\multicolumn{8}{ |c|}{\textsf{\textbf{Experimental Group 1 - 8}}} \\
\hline
\hline
Curved &1.4 &2.1 &2.4 &1.2 &3.2 &3.2 &2.0 &3.1 \\
\hline
Intersected &1.1 &2.2 &1.5 &0.4 &1.8 &2.5 &1.6 &2.7\\
\hline
\end{tabular}
\vspace{-0.4cm}
\end{table}

\begin{figure}[b]
\centering
\includegraphics[width=0.47\textwidth]{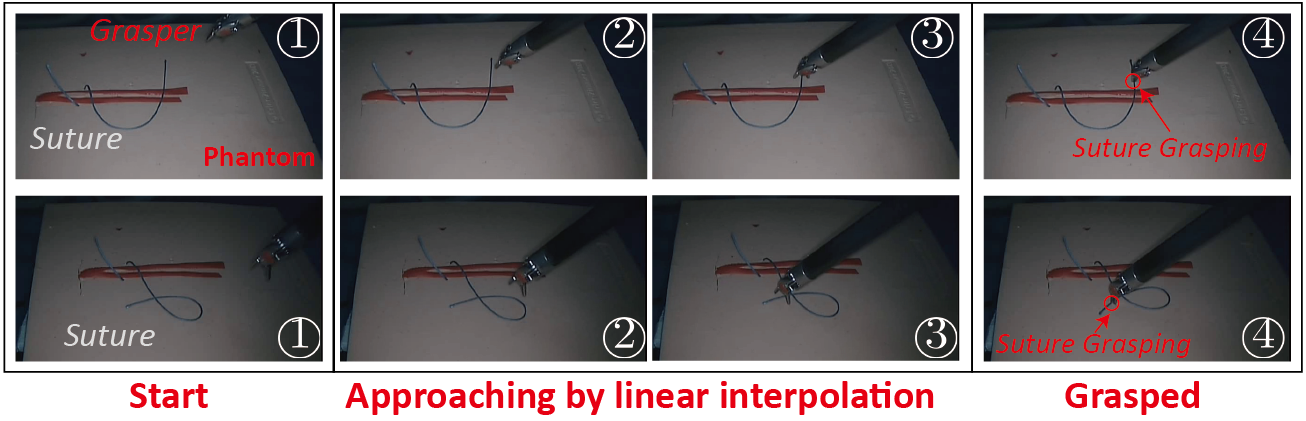}
	\caption{Snapshots of automated grasping of a suture in a dVRK system based on the deep learning-driven framework.}	
	\label{figure_dVRK_grasping}
\end{figure}

The proposed framework was further validated on a dVRK, and two successful trials of this vision-based automated suture's grasping were shown in Fig. \ref{figure_dVRK_grasping}. Summarizing all results above, we can demonstrate the feasibility of our proposed framework to automate the intermediate suture thread grasping step among the suturing, which contributes to a full automation of a robot-assisted surgical knot tying.

\section{Conclusions and Discussions}
\label{section_conclusion}
This paper aims to tackle the technical challenges of automated suture grasping during an interrupted suturing task. A deep learning model, which transfers the information of large legacy surgical data to online acquired images, is leveraged in our framework for suture segment under on-site environmental noises. 
Based on this result, we further propose an algorithm for ordering sequence inference within stereo images, and the 3D coordinates of a designated suture can be thereby obtained by our spatial shape optimizer. 
Utilizing this novel framework, the grasping point can be achieved for an automated grasping manipulation.

We extensively evaluated the segmentation performance comparing our fine-tune model to models that were trained by legacy surgical data and limited online images. Our proposed approach obtained the highest IoU and F1 Scores, which were \(71.1\%\) and \(97.5\%\). 
Besides, typical tests of suture's ordering sequence inference under multiple topologies were listed.
By integrating a UR5 robot with a dual-camera system, experiments concerning approaching key points along the constructed suture were carried out to exploit the overall computational accuracy, and the average errors in 5 testing set were all below \(2.3mm\). 
We also demonstrated the feasibility of the entire framework by simulations in V-REP and by robot experiments in a dVRK system.

By tackling the critical issues in suture's perception, it can be concluded that our framework can provide a reliable guidance to achieve a higher level of assisted task autonomy in a suturing manipulation.

\end{document}